\documentclass[11pt]{article}
\usepackage{coling2020}
\usepackage{times}
\usepackage{url}
\usepackage{latexsym}
\usepackage{enumitem}
\usepackage{multirow}
\usepackage{amsmath}
\usepackage{amsfonts}
\usepackage{graphics}
\usepackage{graphicx}
\usepackage{multicol}
\usepackage{multirow}
\usepackage{xcolor}
\usepackage{subcaption}
\usepackage{booktabs} 
\usepackage{tabularx}
\usepackage[normalem]{ulem}
\useunder{\uline}{\ul}{}

\colingfinalcopy % Uncomment this line for the final submission

\title{Medical Knowledge-enriched Textual Entailment Framework}
\author{
Shweta Yadav$^{\ast}$, Vishal Pallagani$^{\ddagger}$, Amit Sheth$^{\ddagger}$ \\
  $^{\ast}$LHNCBC, U.S. National Library of Medicine, MD, USA  \\
 $^{\ddagger}$ University of South Carolina, SC, USA  \\
  {\tt 
  $^{\ast}$shweta.shweta@nih.gov}, 
  {\tt $^{\ddagger}$\{VISHAL,AMIT\}@sc.edu
  }
  }

\date{}

\begin{document}
\maketitle
\begin{abstract}
One of the cardinal tasks in achieving robust medical question answering systems is textual entailment. The existing approaches make use of an ensemble of pre-trained language models or data augmentation, often to clock higher numbers on the validation metrics. However, two major shortcomings impede higher success in identifying entailment: \textbf{(1)} understanding the focus/intent of the question and \textbf{(2)} ability to utilize the real-world background knowledge to capture the context beyond the sentence. 
% In this paper, we present a novel Knowledge-Enriched Natural Language Inference framework focused on medical domain. The framework utilizes the dual-encoding strategy by jointly modeling the knowledge expanded graph network with the BERT language encoder. The former enables the model to acquire the global semantic context from medical knowledge and later helps to capture local characteristics of medical-entities in the context, complementing each other. 
In this paper, we present a novel Medical Knowledge-Enriched Textual Entailment framework that allows the model to acquire a semantic and global representation of the input medical text with the help of a relevant domain-specific knowledge graph. 
% The framework utilizes the dual-encoding strategy by jointly modeling the traditional document encoder with the knowledge enriched Graph encoder to effectively learn the global context while complementing local characteristics of medical-entities. 
We evaluate our framework on the benchmark \texttt{MEDIQA-RQE} dataset and manifest that the use of knowledge-enriched dual-encoding mechanism help in achieving an absolute improvement of 8.27\% over SOTA language models. We have made the source code available here.\footnote{ \url{https://github.com/VishalPallagani/Medical-Knowledge-enriched-Textual-Entailment}} 
\end{abstract}

\blfootnote{
    
    % % final paper: en-uk version 
    %
    % \hspace{-0.65cm}  % space normally used by the marker
    % This work is licensed under a Creative Commons 
    % Attribution 4.0 International Licence.
    % Licence details:
    % \url{http://creativecommons.org/licenses/by/4.0/}.
    % 
    % % final paper: en-us version 
    %
    % \hspace{-0.65cm}  % space normally used by the marker
    This work is licensed under a Creative Commons 
    Attribution 4.0 International License.
    License details:
    \url{http://creativecommons.org/licenses/by/4.0/}.
}
\section{Introduction}
\label{intro}

The entailment task is similar with natural language inference (NLI), involves identifying the semantic similarity between two natural language texts, premise ($P$) and hypothesis ($H$). The NLI task's effectiveness is crucial for developing a robust natural language understanding system that functions at a human level \cite{MEDIQA2019,abacha2016recognizing,romanov2018lessons}. 
% It also helps in fine-tuning further NLP tasks such as question-answering, information extraction, and text summarization.
Recent literature suggests the use of contemporary language models (LMs) \cite{devlin2018bert,beltagy2019scibert}, often ensembled, to achieve better performance \cite{zhu2019panlp,bhaskar2019sieg,xu2019doubletransfer} on the NLI task. However, our qualitative interpretation of the dataset and results suggests that LMs fails when it comes to textual entailment (TE), despite being the on-demand language model. The limitations belong to two major categories:
\begin{itemize}[nolistsep]
    \item \textbf{Multiple word form:} Medical text offers high degree of variability in the form of synonym and abbreviated words. 
    % The existence of this arbitrariness increases the difficulty level in capturing the semantic relationships between the medical-entities, which often leads to the failure of contemporary LMs in recognizing entailment. 
    The same can be witnessed in Table-\ref{table:intro_example}, where BERT \cite{devlin2018bert} is unable to predict the entailment between $P$ and $H$ as they have different terms - \textit{`Kartagener's Syndrome'} and \textit{`Primary Ciliary Dyskinesia'}, while both are synonym.
    \item \textbf{Focus/Intent understanding:} Given $P$ and $H$, LMs often fails to capture the focus/intent of both the sentences. Table-\ref{table:intro_example} shows an example, where the focus of $P$ and $H$ are misunderstood. It can be seen that $P$ emphasizes the possibility of \textit{`atypical pneumonia'} occurring within a month after treatment, whereas H talks about the possible treatments for the disease. 
    %BERT misclassifies the above as entailment, though the actual label is otherwise.
\end{itemize}
\begin{table}[h]
% \footnotesize
\centering
\resizebox{0.85\linewidth}{!}{
\begin{tabularx}{\linewidth}{>{\hsize=0.1\hsize}X|
                              >{\hsize=0.9\hsize}X}
\toprule
\multirow{2}{*}{\rotatebox{45}{\textbf{entails}}} &
\textbf{Premise:}  I am suffering from \textcolor{blue}{Kartagener's syndrome} and wanted information from you or from Dr. [NAME] for this syndrome. \textcolor{blue}{(About fertility)} and if possible other symptoms. \\
&\textbf{Hypothesis:} \textit{What is \textcolor{blue}{primary ciliary dyskinesia?}}\\
\hline
\multirow{2}{*}{\rotatebox{45}{\textbf{not entails}}}
 &
\textbf{Premise:} What is the \textcolor{red}{possibility of atypical pneumonia occurring again less than a month after treatment?}\\
&\textbf{Hypothesis:} \textit{What are the \textcolor{red}{possible treatments for atypical pneumonia?}}\\
\hline
\end{tabularx}
}
\caption{Examples from the MedQA-RQE dataset, where the text highlighted in \textcolor{blue}{blue} are semantically similar words and \textcolor{red}{red} represents the lexically similar but semantically dissimilar words.}
%A sample example of passages, answer and question triplet where one has to consider all the supporting facts (shown in \textcolor{blue}{blue}) in order to generate the correct question for given answer.}
\label{table:intro_example}
% \vspace{-1.5 em}
\end{table}
% \vspace{-1em}

These findings indicate that existing LMs lack semantic interpretation of the input, which is crucial in the inferencing tasks. In this paper, we deal with the question:\\
\indent \textit{Does the medical textual entailment task benefit from the external domain knowledge to distinguish semantically identical medical sentences to recognize entailment? }\\
\indent To address this question and above-mentioned limitations, this paper presents a novel framework for recognizing textual entailment, \texttt{Sem-KGN}: \textbf{Sem}antic \textbf{K}nowledge-enriched \textbf{G}raph \textbf{N}etwork that explores the domain-specific knowledge to enhance the semantic interpretability in the LMs. The proposed method devise a dual-encoding mechanism to enrich the classical document encoding (obtained from BERT) with the knowledge-enriched graph encoder. Specifically, our method builds a heterogeneous dictionary graph of the given $P$ and $H$ to encode the global context, while BERT's proficiency lies in capturing the local contextual information. Rather than constructing graphs solely based on the triples (subject, object, predicate), more \textit{semantic units} are introduced into the graph as additional nodes to enrich the relations between the entities. These additional nodes add medical-entities centered factual information which are generated by expanding the medical knowledge graphs (KGs) such as UMLS \cite{bodenreider2004unified}, SNOMED-CT \cite{donnelly2006snomed} and ICD10 \cite{quan2005coding}. The medical entities present in $P$ and $H$ are used to query the related information, \textit{`diseases/syndromes'}, \textit{dosage}, \textit{`side-effects'}, and \textit{`drug-interaction'} from the mentioned KGs. Finally, Graph Convolutional Network (GCN) \cite{kipf2017semi} is employed to generate the graph encoding, augmented with the regular document encoder, and later fused through a multi-headed attention layer to support entailment.\\
%In essence the motivation of overcoming the absence of real-world medical knowledge and addressing the issue of understanding the context is achieved by proposed \texttt{Sem-KGN}. \\
\textbf{Contributions: (i)} Proposed medical-TE framework, by utilizing domain-specific medical KGs to encode the global and semantic information of the premise and hypothesis, \textbf{(ii)} Exploited the capabilities of semantic units in a graph network to encode medical-entities centered factual information for recognizing textual entailment, and \textbf{(iii)} Evaluated the effectiveness of the proposed method over state-of-the-art language models and knowledge-infused baseline methods on the benchmark MEDIQA-RQE dataset.
\vspace{-0.5 em}
\paragraph{Related Work :}
% \vspace{-0.5 em}
The development of annotated TE and NLI medical datasets \cite{abacha2015semantic,MEDIQA2019,abacha2016recognizing,romanov2018lessons} and a variety of pre-trained language models has led to a rise of extensive ongoing research in this field. Majority of the systems developed for the TE task adopts the multi-task learning (MTL) framework \cite{zhu2019panlp,bhaskar2019sieg,kumar2019dr,zhou2019dut,xu2019doubletransfer}, ensemble method \cite{sharma2019iit}, and transfer learning \cite{bhaskar2019sieg} for achieving better accuracy. 
\newcite{xu2019doubletransfer} employed the MTL approach \cite{liu2019improving,yadav2018multi,yadav2019unified,yadav2020relation} in TE task to learn from the auxiliary tasks of question answering (QA) and NLI.
The best performing system at MedQA 2019-RQE shared task \cite{zhu2019panlp} utilized the MTL approach to learn from intermediate NLI task. Further they used knowledge distillation approach to condense the information obtained from various models and transfer it into an single model. 
% The early approaches for building an entailment classifier solely considered the textual information as an input for deep learning models \cite{yang2019enhancing}. Later on, attention mechanism \cite{vaswani2017attention} has been widely adopted for better understanding of local information in text. A multi-task \cite{liu2019improving} approach of language models is further employed to achieve state-of-the art results \cite{xu2019doubletransfer}. 
% MTL based approaches, aims to learn global context for entailment not available in the local context of a single task and possibly ``learnable" by training across multiple tasks. Such global context is statistically derivable from the inputs. In the absence of external knowledge, given enough data across various contexts, even if this is considered a reasonable assumption, the data required to form these abstract global contextual connections is likely to be huge in volume. However, our method uses already existing wealth of explicit semantic knowledge not requiring gathering of noise free, sizable amounts of data in-order for this assumption to work.
% Few of the works \cite{8970629,wang2019improving,khot2018scitail} have explored the usage of the background knowledge or medical KGs in extracting information for the entailment task.  
Few of the works \cite{wang2019improving,khot2018scitail} have explored the usage of the background knowledge or medical KGs \cite{kumar2019dr,bhaskar2019sieg} in extracting information for the entailment task. 
% In the related task of medical-NLI, some of the prominent studies have explored knowledge-enriched co-attention \cite{chen-etal-2018-neural}, concept definition from UMLS \cite{lu2019incorporating}, and knowledge graph embedding \cite{sharma2019incorporating}. 
However, the consideration of building a vocabulary graph from the textual and later enriching them with information from the medical KGs is still an unexplored territory.

\begin{figure}[h]
\includegraphics[width=\textwidth,height=7cm,keepaspectratio]{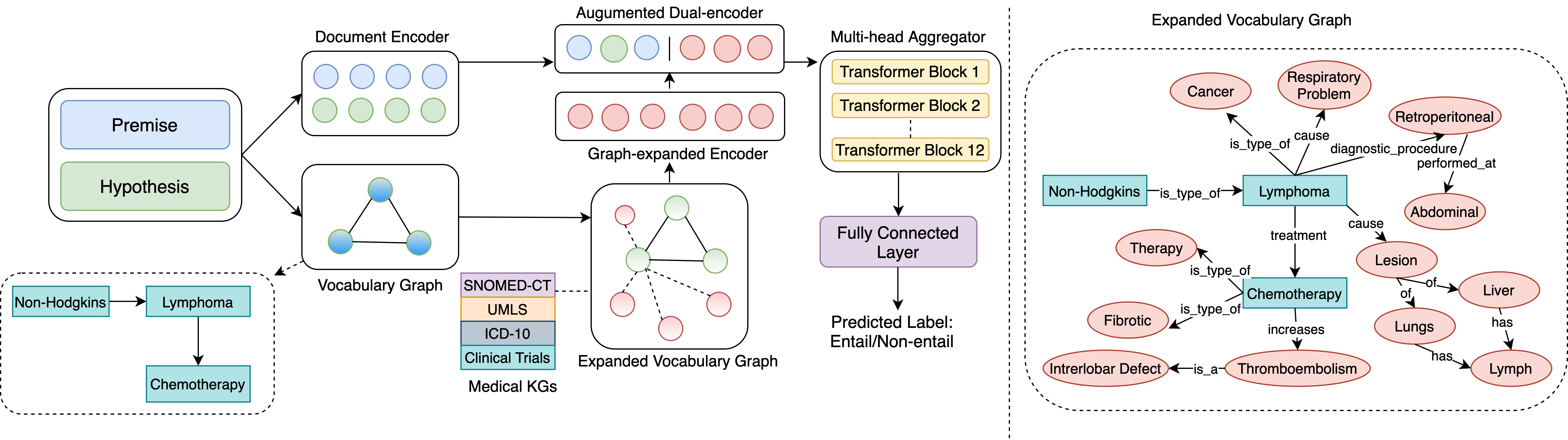}
\caption{Architecture of proposed methodology}
\label{architecture}
\end{figure}
\vspace{-0.5 em}
\section{Proposed Approach}
The overall architecture of the proposed \texttt{Sem-KGN} is illustrated in Fig.-\ref{architecture}. 
% The entire pipeline can be broken down into six major phases with the sole purpose of effectively recognizing entailment. Phase 1 deals with the generation of the word embeddings from P and H using BERT, whose novelty lies in applying bidirectional training of a Transformer. The succeeding phase involves generation of the vocabulary graph considering P and H. The medical entities are identified in the vocabulary graph in Phase 3 and enriched with information from KGs such as SNOMED-CT, ICD 10, UMLS, and Clinical Trials. GCN is used to compute the graph embeddings which are augmented with BERT embeddings in Phase 4. The obtained embeddings are passed through 12 layers of self-attention mechanism in Phase 5. The final phase consists of a fully connected layer for classifying P and H as entailment or not.
The rest of the section elaborates \texttt{Sem-KGN} in detail.
% The entire pipeline can be broken down into six major phases with the sole purpose of effectively recognizing entailment. Phase 1 deals with the generation of the word embeddings from P and H using BERT, whose novelty lies in applying bidirectional training of a Transformer. The succeeding phase involves generation of the vocabulary graph considering P and H. The medical entities are identified in the vocabulary graph in Phase 3 and enriched with information from KGs such as SNOMED-CT, ICD 10, UMLS, and Clinical Trials. GCN is used to compute the graph embeddings which are augmented with BERT embeddings in Phase 4. The obtained embeddings are passed through 12 layers of self-attention mechanism in Phase 5. The final phase consists of a fully connected layer for classifying P and H as entailment or not. The rest of the section elaborates Sem-KGN in detail.
% \vspace{0.5 em}
\subsection{Document Encoder}
Given a $n_P$ words premise $P = \{w_1^P, w_2^P, \ldots, w_{n_P}^P\}$ and $n_H$ words hypothesis $H = \{w_1^H, w_2^H, \ldots, w_{n_H}^H\}$. The document encoder is responsible to capture basic lexical and syntactic information from the premise and hypothesis in the form of local context information. We employed the BERT model to serve as the document encoder in our proposed \texttt{Sem-KGN} framework.
% To be specific, given the token sequence (say $n$ words) of $P$ and $H$ the document encoder firstly sums the token embedding, segment embedding, positional embedding \cite{devlin2018bert} for each token to compute its input embedding. 
% The input embedding pass to the $L$ layers Transformer architecture \cite{vaswani2017attention} to compute the $d$-dimensional contextual representations $B \in \mathbb{R}^{n \times d}$. 
Formally, the local context features are computed as
$ d_1, d_2, \ldots, d_n = \texttt{Document-Encoder} (w_1, w_2, \ldots, w_n)$.
% $\{d_1, d_2, \ldots, d_n\}$ are computed as follows:
% \begin{equation}
% \label{doc-encoder}
%     d_1, d_2, \ldots, d_n = \texttt{Document-Encoder} (w_1, w_2, \ldots, w_n)
% \end{equation}

% BERT provides a major breakthrough in the language models by training the transformers in a bidirectional fashion. The transformer encoder assimilates the entire sequence of words at once, allowing the model to learn the context of every word in the sentence. Thus, the BERT embeddings help in understanding the local context of the sentence. Phase 1 constitutes the beginning of the ExBERT pipeline, which later takes into the consideration of a knowledge-enriched vocabulary graph to capture global contextual understanding of a sentence.

% BERT provides a major breakthrough in the language models by training the transformers in a bidirectional fashion. The transformer encoder assimilates the entire sequence of words at once, allowing the model to learn the context of every word in the sentence. Thus, the BERT embeddings help in understanding the local context of the sentence. Phase 1 constitutes the beginning of the ExBERT pipeline, which later takes into the consideration of a knowledge-enriched vocabulary graph to capture global contextual understanding of a sentence.
\subsection{Knowledge-enriched Graph Encoder}
\textbf{Vocabulary Graph Construction:}
We first construct a dictionary based on all the unique words in the training dataset. Thereafter, we build a graph $G=(V,E)$ based on the word co-occurrence information in the dictionary. Instead of building graph based on a given $P$ and $H$, we were motivated by the work of \cite{lu2020vgcn} to build graph by considering all the lexicon in the dataset, which aims at encoding the global information of the particular domain (in our case medical domain). The nodes of the graph $G$ are words in the dictionary, the edge between two nodes $w_i$ and $w_j$ is determine by the normalized point-wise mutual information (NPMI) \cite{bouma2009normalized}. 
\begin{equation}\label{eq:npmi}
\footnotesize
\texttt{NPMI}(w_i, w_j) = - ln \frac{p(w_i, w_j)}{p(w_i)p(w_j)} \frac{1}{ln~p(w_i,w_j)}
\end{equation}
where  $p(w_i, w_j) = \frac{\#s(w_i, w_j)}{\#W}$, $p(i) = \frac{\#s(w_i)}{\#W}$, $\#s(.)$ is the number of sliding windows containing a word or a pair of words, and $\#W$ is the total number of sliding windows.
We make an edge between the nodes if the value of $\texttt{NPMI}(w_i,w_j)$ exceed a particular threshold value.
\paragraph{Graph-expansion with Medical Knowledge:} We expand the existing graph $G$ with the additional nodes and corresponding edges to form a graph $G^*=(V \cup \overline{V}, E \cup \overline{E})$. Towards this, first we extracted the medical entities by exploiting the entity recognition model\footnote{\url{https://go.aws/37SD7ae}} trained on MedMentions dataset \cite{mohan2019medmentions}, from a given pair $P$ and $H$. Once the medical entities are identified, KGs such as SNOMED-CT, ICD-10, UMLS and Clinical Trials are exploited to extract the information of two-hop connected \textit{`diseases/syndromes'}, \textit{`dosage'}, \textit{`side effects'}, and \textit{`drug-interaction'} type medical-concepts. 
%We expand each extracted medical entity with their appropriate entities with the help of ------. 
The final expanded entities are the additional nodes $\overline{V}$ which act as the \textit{semantic unit} to capture the domain-specific relationship (e.g., \textit{treats}, \textit{caused by}) and hierarchical relations (e.g., \textit{is a}) between medical-concepts. \\
% The extracted nodes and the relations are used to form an additional edge $\overline{E}$ with the corresponding medical term.\\
%In order to get the edges between them, we identify the relation between any two medical entities by performing the -----.
\textbf{Graph-expanded Knowledge Encoding:} The \textbf{G}raph-e\textbf{x}panded \textbf{K}nowledge \textbf{Encoder} (\texttt{GxK Encoder}) computes the representation of each node from graph $G^*$. We are interested to compute the representation for each extracted entities ($e_1, e_2 \ldots e_m$) for given pair of $P$ and $H$.
% , premise $P$ and hypothesis $H$. It computes the representation of each node in the graph $G^*$
Formally, we get $ h_1, h_2, \ldots, h_m = \texttt{GxK-Encoder}(G^*, e_1, e_2, \ldots, e_m)$\\
% \begin{equation}
% \label{graph-encoder}
%     h_1, h_2, \ldots, h_m = \texttt{GxK-Encoder}(G^*, P, H) = \texttt{GxK-Encoder}(G^*, e_1, e_2, \ldots, e_m)
% \end{equation}
\indent We model it using the 2-layer GCN architecture. For the entities, we first make a input matrix  $M \in \mathbb{R}^{m \times |V|}$, where row of the matrix $M$ is the one-hot vector of length of $|V|$ (size of dictionary). Given the adjacency matrix $D$ of expanded graph $G^*$ a and input matrix $M$, a single layer of graph convolution is computed as follows:
\begin{equation}
\footnotesize
  H^{(1)} = relu(MDW^{(1)}), ~ \text{where} ~ D \in \mathbb{R}^{|V| \times |V|} ~ \text{and} ~  W^{(1)} \in \mathbb{R}^{|V| \times d^{}}
  \label{eq_gcn}
\end{equation}
where $H^{(1)} \in \mathbb{R}^{m \times d}$ is matrix which rows are the node features in. For a given input $M$, we are interested to captures the part of the graph from $D$ by multiplying them together as $MD$. The feature at a given node $w_i$ computed by the interaction between all the neighbouring nodes as $h^{(l+1)}_{w_i} = relu \left( \sum_{j} \frac{1}{c_{ij}}h^{(l)}_{w_j}W^{(l)} \right) \,$ where $c_{ij}$ is the normalization constant \cite{kipf2017semi}. The second layer convolution is obtained as $ H^{(2)} = relu(H^{(1)}W^{(2)})$, where $W^{(2)} \in \mathbb{R}^{d \times d^{}}$
\subsection{Multi-headed Aggregator} We fuse the information from \texttt{Document-Encoder} and \texttt{GxK-Encoder} using the multi-headed self-attention \cite{vaswani2017attention}. Our aims to utilize the best of both the worlds. 
% We capture the local context information from document encoder and global and external knowledge from knowledge encoder. 
We form a single feature sequence by augmenting both the encoder representations obtained from both the encoders, with additional \texttt{[CLS]} token representation. By applying the self-attention on the augmented encoding sequence, we facilitate the network to attend the useful information across the individual encoding. We model our \textbf{M}ulti-\textbf{H}eaded \textbf{Aggregator} (\texttt{MH-Aggregator}) using the $12$ layers of Transformer-block with $16$ heads. 
% Formally, a \texttt{Transformer-block} $F^{l}$ at layer $l$ comprised of multi-headed self-attention attention followed by the point-wise feed-forward layer.
We use the last layer output of the aggregator and consider the \texttt{[CLS]} token representation as the final representation $F \in \mathbb{R}^{d}$ of given $P$ and $H$ pair. We employ a feed-forward layer to classify a pair of premise $P$ and hypothesis $H$ into the corresponding \textit{`entail'} or \textit{`non-entail'} classes.
\begin{equation}
\footnotesize
\label{mh-aggregator}
\begin{aligned}
f_1, f_2, \ldots, f_{n+m} &= \texttt{MH-Aggregator}(d_1, d_2, \ldots, d_n, h_1, h_2, \ldots, h_m) \\
prob ({class=entail}|P, H, \theta) &= exp(W_{entail}^{T}F+b)/ \sum_j{exp(W_{j}^{T}F+b)} 
\end{aligned}
\end{equation}

\begin{minipage}{\textwidth}
  \begin{minipage}[b]{0.5\textwidth}
\resizebox{\linewidth}{!}{
\begin{tabular}{l|l|llll}
\hline
\multicolumn{2}{c|}{\textbf{Models}} & \textbf{Accuracy} & \textbf{Precision} & \textbf{Recall} & \textbf{F1-Score} \\ \hline
\multirow{3}{*}{\textbf{Baseline 1}} & BERT & $47.90$ & $46.16$ & $48.26$ & $47.18$ \\ 
 & BioBERT & $50.14$ & $46.28$ & $48.97$ & $47.58$ \\ 
 & ClinicalBERT & $49.60$ & $48.79$ & $49.56$ & $49.17$ \\ \cline{1-2} \hline
\multirow{3}{*}{\textbf{Baseline 2}} & BERT + KI & $49.56$ & $46.06$ & $48.69$ & $47.33$\\ 
 & BioBERT + KI & $51.15$ & $48.56$ & $49.56$ & $49.05$\\ 
 & ClinicalBERT + KI & $50.14$ & $50.00$ & $50.00$ & $50.00$\\ \cline{1-2} \hline
 \textbf{\cite{zhu2019panlp}} & BERT + linear projection & \textit{\textbf{$51.30$}} & \textit{\textbf{$51.53$}} & \textit{\textbf{$51.30$}} & \textit{\textbf{$49.45$}} \\ \cline{1-2} \hline
%   \textbf{\cite{zhu2019panlp}} & BERT + linear projection-DEV & \textit{\mathbf{$77.81$}} & \textit{\textbf{$78.35$}} & \textit{\textbf{$78.86$}} & \textit{\textbf{$77.77$}} \\ \cline{1-2} \hline
%   \textbf{\cite{zhu2019panlp}} & BERT-DEV & \textit{\mathbf{$78.81$}} & \textit{\textbf{$78.41$}} & \textit{\textbf{$78.84$}} & \textit{\textbf{$78.54$}} \\ \cline{1-2} \hline
\textbf{Proposed Model} & \texttt{Sem-KGN} & $\textbf{56.17}$ & $\textbf{63.18}$ & $\textbf{56.18}$ & $\textbf{59.47}$\\ \cline{1-2} \hline
\end{tabular}
}
\captionof{table}{Experimental results of our proposed model (\texttt{Sem-KGN}) and the baseline methods on the official test set.}
\label{results}
\end{minipage}
 \begin{minipage}[b]{0.45\textwidth}
\resizebox{\linewidth}{!}{
\begin{tabular}{l|cccc}
\hline
\textbf{Models Components} & \textbf{Accuracy} & \textbf{Precision} & \textbf{Recall} & \textbf{F1-Score} \\ \hline
\texttt{Sem-KGN} & $56.17$ & $63.18$ & $56.18$ & $59.47$ \\  \hline
\begin{tabular}[c]{@{}l@{}}
\textbf{(-)} Knowledge-enriched \\ Graph Encoder
\end{tabular} & $47.90$ ($8.27\downarrow$) & $46.16$ ($17.02\downarrow$) & $48.26$ ($7.90\downarrow$) & $47.18$ ($12.29\downarrow$)\\ \hline
\begin{tabular}[c]{@{}l@{}}
\textbf{(-)} Medical \\Knowledge-graph
\end{tabular} & $50.65$ ($5.52\downarrow$) & $62.27$ ($0.91\downarrow$) & $50.66$ ($5.52\downarrow$) & $55.86$ ($3.61\downarrow$)\\ \hline 
\end{tabular}
}
\captionof{table}{Ablation study showing the role of each component in the model. The values within the bracket show the absolute decrements by removing the component.}

\label{ablation-study}
\end{minipage}
\end{minipage}
\section{Experimental Results and Analysis}
\paragraph{Dataset and Metrics:}
We used widely adopted benchmark entailment dataset, MEDIQA-RQE created by \cite{abacha2016recognizing}, released in the BioNLP 2019 shared task. 
The dataset is derived from consumer health questions (CHQs) and frequently asked questions (FAQs) from the U.S. National Library of Medicine and National Institute of Health respectively. The training and validation set consists of total $8890$ CHQ and FAQ pairs with the entail label of $4784$ instances. The test set consist of $230$ pairs with $115$ entail labels. We used official evaluation metrics (Accuracy) to evaluate our model. Additionally, we also provided the Precision, Recall, and F1-Scores for the evaluation. 
\paragraph{Implementation Details:} We have chosen models' hyper-parameters empirically on the validation set. The base-uncased version of BERT\footnote{\url{https://bit.ly/2ZajZR1}} of hidden size $768$ with a max sequence length of $200$ ($160$ for $P$ and $40$ for $H$) is used in all experiments reported in the paper. The size of dictionary to create the dictionary matrix was $30000$. The dimension of  The threshold of NPMI is set to $0.3$ to obtain meaningful relation between words. The last layer's hidden size of the graph-expanded knowledge encoder is set to $16$. We use the Adam optimiser \cite{kingma2014adam} for parameters update after every epoch of training. We set the  standard batch size of $16$, and trained for $5$ epochs with a dropout rate of $0.2$ and $2e-5$ learning rate in all the experimental results reported in this work.\\
% \subsection{Experimental Setup}
% \paragraph{Dataset and Metrics:}
% We used widely adopted benchmark entailment dataset, MEDIQA-RQE created by \cite{abacha2016recognizing}, released in the BioNLP 2019 shared task. 
% The dataset is derived from consumer health questions (CHQs) and frequently asked questions (FAQs) from the U.S. National Library of Medicine and National Institute of Health respectively. The training and validation set consists of total $8890$ CHQ and FAQ pairs with the entail label of $4784$ instances. The test set consist of $230$ pairs with $115$ entail labels. We used official evaluation metrics (Accuracy) to evaluate our model. Additionally, we also provided the Precision, Recall, and F1-Scores for the evaluation. 
% \paragraph{Implementation Details:} We have chosen models' hyper-parameters empirically on the validation set. The base-uncased version of BERT\footnote{\url{https://bit.ly/2ZajZR1}} of hidden size $768$ with a max sequence length of $200$ ($--$ for $P$ and $--$ for $H$) is used in all experiments reported in the paper. The size of dictionary to create the dictionary matrix was $---$. The dimension of  The threshold of NPMI is set to $0.3$ to obtain meaningful relation between words. The last layer's hidden size of the graph-expanded knowledge encoder is set to $16$. We use the Adam optimiser \cite{} for parameters update after every epoch of training. We set the  standard batch size of $16$, and trained for $5$ epochs with a dropout rate of $0.2$ and $2e-5$ learning rate in all the experimental results reported in this work. 
\textbf{Baseline Models:}
To show the effectiveness of \texttt{Sem-KGN}, we adopted following competitive baseline models:\\
% We used widely adopted benchmark entailment dataset, MEDIQA-RQE created by \cite{abacha2016recognizing}, released in the BioNLP 2019 shared task and evaluated our model on official evaluation metrics (Accuracy). To show the effectiveness of \textit{Sem-KGN}, we adopted following competitive baseline models:
\textbf{1. Language Models (LMs):} We utilized the SOTA LMs (\textit{BERT}) as well as the LMs adapted for the clinical (\textit{ClinicalBERT}) and medical domain (\textit{BioBERT}), fine-tuned for MEDIQA-RQE task.\\
\textbf{2. Knowledge-Infused Language Models (+KI):} These baselines model works on the principle of \textit{shallow knowledge infused learning}, where we provided the knowledge about the medical entities at the token level to the LMs. The intuition behind using this baseline was to understand at what layer if knowledge is integrated into the LMs, it is going to be beneficial.  
% \begin{enumerate}[nolistsep]
%     \item\textbf{ Language Models (LMs):} We utilized the SOTA LMs (\textit{BERT}) as well as the LMs adapted for the clinical (\textit{ClinicalBERT}) and medical domain (\textit{BioBERT}), fine-tuned for MedQA task.
%     \item\textbf{Knowledge-Infused Language Models (+KI):} These baselines model works on the principle of \textit{shallow knowledge infused learning}, where we provided the knowledge about the medical entities at the token level to the LMs. The intuition behind using this baseline was to understand at what layer if knowledge is integrated into the LMs, it is going to be beneficial.  
% \end{enumerate}
%We also introduced another series of \textit{knowledge-infused baselines model (\textbf{+KI})}. 
\paragraph{Results:}
Table-\ref{results} provides an overview of the results, which demonstrates that \texttt{Sem-KGN}, equipped with KGs enriched graph encoding performs the best over all the baselines model. A considerable increase in the accuracy of $8.27$\% can be observed over vanilla BERT model in comparison to proposed \texttt{Sem-KGN}. The similar set of improvement (over $6$\%) can be observed with BioBERT and ClinicalBERT. We also observed the power of basic shallow (+KI) over the vanilla LMs showing the absolute improvement of $1.5$\%. Further, in comparison to baseline 2, \texttt{Sem-KGN} achieved the average increment of $6$\% over all the knowledge-infused LMs.
Finally, from our ablation study (\textit{c.f.} Table-\ref{ablation-study}), it can be noticed that enriching the graph encoder with the domain knowledge assists in the entailment task.
The results conclude two important claims: \textbf{(1)} the modularity of the knowledge infusion process that can be combined with any LMs is witnessed, and \textbf{(2)} \texttt{Sem-KGN} proves its effectiveness in having a local as well global understanding of the premise and hypothesis. We also compare the results against the best system \cite{zhu2019panlp} at MEDIQA-RQE task that was based on the ensemble of LMs. However, to have a fair comparison and understand the role of our Knowledge-enriched graph encoder, we only utilize their model that have introduced ``linear projection" over BERT. The proposed \texttt{Sem-KGN} has outperform the ``linear projection" mechanism described in \newcite{zhu2019panlp}. \\
\textbf{Analysis:}   
Table-\ref{tab:analysis} depicts the qualitative analysis of \texttt{Sem-KGN}, w.r.t baseline models on $50$ randomly sampled DEV set. The first entry shows the effectiveness of Sem-KGN in being able to understand the focus of the premise and the hypothesis which is achieved by the dual encoding mechanism. The second entry in the table affirms the importance of domain-specific KGs in assimilating medical information. \\
\textbf{Error Analysis:} Table-\ref{tab:analysis} entry $3$ and $4$ shows the leading cause of the error in the proposed model. We found that major misclassification occurred when the premise was complex and have a multiple questions. We also observed in some cases when there is high ambiguity between premise and hypothesis, model fail to recognize the correct label. For e.g., in the Table-\ref{tab:analysis} entry 4, with the presence of term \textit{`hyperthyroidism'} and \textit{`diagnosed'} both in premise and hypothesis it is very difficult to recognize the true label: not-entailed.

\begin{table*}[]
\centering
\resizebox{\textwidth}{!}{%
\begin{tabular}{c|l|l|l|l|l}
\hline
\multicolumn{1}{l|}{\multirow{2}{*}{\textbf{Sample Examples and Error Type}}} & \multicolumn{1}{c|}{\multirow{2}{*}{\textbf{Premise}}} & \multicolumn{1}{c|}{\multirow{2}{*}{\textbf{Hypothesis}}} & \multirow{2}{*}{\textbf{True Label}} & \multirow{2}{*}{\textbf{BioBERT+ KI}} & \multirow{2}{*}{\textbf{Sem-KGN}} \\
\multicolumn{1}{l|}{} & \multicolumn{1}{c|}{} & \multicolumn{1}{c|}{} &  &  &  \\ \hline
\textbf{Example 1} & \textit{\begin{tabular}[c]{@{}l@{}}Can you mail me patient information about Glaucoma, I was recently \\ diagnosed and want to learn all I can about the disease.\end{tabular}} & \textit{How is glaucoma diagnosed ?} & Non-entailed & Non-entailed & Non-entailed \\ \hline
\textbf{Example 2} & \textit{\begin{tabular}[c]{@{}l@{}}I was writing to inquire about more information regarding the \\ diagnosis of OI.  We have family members who are in the process of \\ waiting for genetic testing to come back but are under allegations of \\ child abuse. Is there any information that may be helpful to us?\end{tabular}} & \textit{How to diagnose Osteogenesis Imperfecta ?} & Entailed & Non-entailed & Entailed \\ \hline
\textbf{\begin{tabular}[c]{@{}c@{}}Error Type 1:\\ Complex and Long Question\end{tabular}} & \textit{\begin{tabular}[c]{@{}l@{}}Hello, my dad, 68 years old, has gastritis, it did ache occasionally \\ over the last several years. The other day, he went to hospital to have\\ medical check-up with endoscopic ultrasonography, and found GIST \\ with about 1cm in size. Dr. told him that he may consider surgery or \\ not, it is up to him. What are we supposed to do?\end{tabular}} & \textit{How is an endoscopic ultrasound performed ?} & Non-entailed & Entailed & Non-entailed \\ \hline
\textbf{\begin{tabular}[c]{@{}c@{}}Error Type 2:\\ Ambiguous Question\end{tabular}} & \textit{\begin{tabular}[c]{@{}l@{}}Can you please send me as much information as possible on ``hypo-\\ thyroidism”. I was recently diagnosed with the disease and I am\\ struggling to figure out what it is and how I got it.\end{tabular}} & \textit{How is Hypothyroidism diagnosed ?} & Non-entailed & Entailed & Non-entailed \\ \hline
\end{tabular}
}
\caption{Qualitative and error analysis of our proposed model (\texttt{Sem-KGN}) with the best baseline model.}
\vspace{-0.5 em}
\label{tab:analysis}
\end{table*}

\section{Conclusion}
In this paper, we proposed a framework \texttt{Sem-KGN} to recognize medical textual entailment. Our framework utilized the local context from BERT based document encoder and global context by expanding the vocabulary graph with the medical entities obtained from the medical knowledge-bases. 
% We shows the novel effective way to compute the dual encoding along with the efficient aggregator scheme to fuse the multiple encoding. 
We present an efficient aggregator scheme to fuse the multiple encoding. 
The proposed \texttt{Sem-KGN} framework outperformed the competent pre-trained language model and  knowledge-graph enabled language model architectures with fair margin. In future, we plan to explore multi-domain knowledge-graph and efficient graph embedding based techniques for medical textual entailment. 

% This paper presents a systematic approach for knowledge-enriched semantic graphs for the recognition of entailment which plays a crucial role in many NLP applications. The utilisation of KGs to expand the graphs, later to be augmented with BERT embeddings shows significant improvement in the model performance. The analysis of Sem-KGN confirms that the afore-mentioned shortcomings have been overcome, thus leading to a better NLI model. Furthermore, the Sem-KGN approach is modular, which offers the flexibility to be used with any KG and language model. The future work involves exploring the usage of a semantic parser like OpenIE and multi-domain KGs like ConceptNet+UMLS to build the extended vocabulary graphs. Additionally, it would be interesting to see the performance of Sem-KGN on question-answering datasets.
\bibliographystyle{coling}
\bibliography{coling2020}

\end{document}